\documentclass[10pt,twocolumn,letterpaper]{article}

\usepackage{cvpr}
\usepackage{times}
\usepackage{epsfig}
\usepackage{graphicx}
\usepackage{amsmath}
\usepackage{amssymb}
\usepackage{caption}
\usepackage{stmaryrd}
\usepackage{mwe}
\usepackage{authblk}
\usepackage{booktabs}
\usepackage{subfig}
\usepackage{float}
\usepackage[numbers,sort,compress]{natbib}

\usepackage[breaklinks=true,bookmarks=false]{hyperref}

\cvprfinalcopy 

\def\cvprPaperID{****} 

\begin{document}

\title{Towards Lifelong Self-Supervision For Unpaired Image-to-Image Translation}

\renewcommand\Authands{, }
\author{Victor Schmidt\thanks{For correspondence : \texttt{schmidtv@mila.quebec}\\
Work under review for CVPR 2020's Continual Vision Workshop.\\
Code: \url{https://github.com/vict0rsch/LiSS}}}
\author{\ Makesh Narsimhan Sreedhar}
\author{\ Mostafa ElAraby}
\author{\ Irina Rish}

\affil{Mila\\
Universite de Montreal}

\maketitle

\begin{abstract}
 Unpaired Image-to-Image Translation (I2IT) tasks often suffer from lack of data, a problem which self-supervised learning (SSL) has recently been very popular and successful at tackling. Leveraging auxiliary tasks such as rotation prediction or generative colorization, SSL can produce better and more robust representations in a low data regime. Training such tasks along an I2IT task is however computationally intractable as model size and the number of task grow. On the other hand, learning sequentially could incur catastrophic forgetting of previously learned tasks. To alleviate this, we introduce Lifelong Self-Supervision (LiSS) as a way to pre-train an I2IT model (e.g., CycleGAN) on a set of self-supervised auxiliary tasks. By keeping an exponential moving average of past encoders and distilling the accumulated knowledge, we are able to maintain the network’s validation performance on a number of tasks without any form of replay, parameter isolation or retraining techniques typically used in continual learning. We show that models trained with LiSS perform better on past tasks, while also being more robust than the CycleGAN baseline to color bias and entity entanglement (when two entities are very close).
\end{abstract}

\section{Introduction}

\subsection{Motivation}

In recent years generative unsupervised image-to-image translation (I2IT) has gained tremendous popularity, enabling style transfer~\citep{cyclegan} and domain adaptation \citep{cycada}, raising awareness about wars~\citep{deepempathy} and Climate Change~\citep{vicc} and even helping model cloud reflectance fields~\citep{aicd}. I2IT has become a classical problem in computer vision which involves learning a conditional generative mapping from a source domain $\mathcal{X}$ to a target domain $\mathcal{Y}$. For example, generating an image $\hat{y}$ of a zebra conditioned on an image $x$ of a horse. Obviously, there is no ground-truth data for this transformation and we cannot therefore leverage pairs $(x, y)$ to learn this generative mapping. This is the challenge that unpaired I2IT addresses.

One of the main limitations of the I2IT task is that data is often scarce and hard to acquire~\citep{vicc, cyclegan, lee2018diverse}. To overcome this difficulty, self-supervised learning (SSL) appears to be a promising approach. In SSL, a model is trained on an {\em auxiliary task} (e.g., image rotations prediction) that leverages unsupervised data in order to obtain better representations which can help a downstream task learn faster when few (labeled) samples are available~\citep{jing2019self-supervised}.
Given the variety of such potential auxiliary tasks, one could hope to jointly train many of them along with the main task, thereby improving the performance on the latter. However, this may be impractical in the context of I2IT since  the  models are typically quite large, making parallel training of self-supervised and translation tasks computationally intractable. On the other hand, any form of sequential learning  may result into catastrophic forgetting~\citep{FRENCH1999128} and counter the benefits of SSL.  In this paper,  we therefore investigate how {\em continual learning}, a set of approaches designed to make sequential learning across multiple tasks robust to catastrophic forgetting, can be used to enable self-supervised pre-training of unpaired I2IT models.

We show that self-supervised auxiliary tasks improve CycleGAN's translations with more semantic representations and that distillation~\citep{hinton, zhai2019lifelong} retains the knowledge acquired while pre-training the networks. For easier reference, we call this framework "Lifelong Self-Supervision", or \textit{LiSS}, and show its results on CycleGAN's performance in Section \ref{sec:experiments}.

\subsection{Related Work}

Generative Adversarial Networks (GANs)~\citep{gangoodfellow} have had tremendous success in generating realistic and diverse images~\citep{karras2019analyzing, YiLLR19, brock2018large}. Generative I2IT approaches often leverage GANs to align the distributions of the source and target domains and produce realistic translations~\citep{huang2018multimodal}. In their seminal work, \citet{isola2016image} proposed a principled framework for I2IT by introducing a weighted average of a conditional GAN loss, along with an $L_1$ reconstruction loss to leverage pairs (for instance edges $\leftrightarrow$ photos or labels $\leftrightarrow$ facade). To address the setting where pairs are not available, \citet{cyclegan} introduced the cycle-consistency loss which uses two networks to symmetrically model source$\rightarrow$target and source$\leftarrow$target.  The cycle-consistency induces a type of self-supervision by enforcing a reconstruction loss when an image goes through one network and then the other. Many attempts have since been made to improve the diversity \citep{huang2018multimodal, lee2018diverse} and semantic consistency \citep{mo2018instagan, mejjati2018unsupervised} of CycleGAN-inspired I2IT models by leveraging an encoder-decoder view of the models. We keep with the CycleGAN encoder-decoder model, and use self-supervision to encourage the encoder to encode meaningful features for the decoder to work with (see section \ref{sec:approach} for more details).\\

Self-supervised learning tries to leverage the information already present in samples to produce data-efficient representations. This is often measured by pre-training a standard supervised network on an auxiliary (or \textit{pretext}) task, and then measuring its performance on the actual dataset on a fixed, low budget of labels~\citep{jing2019self-supervised}. Though not new~\citep{de1994learning}, it has gained a lot of popularity in the deep learning era with important successes in language modeling \citep{pennington2014glove, devlin2018bert, howard2018universal} speech recognition~\citep{ravanelli2020multitask}, medical imaging~\citep{raghu2019transfusion} and computer vision in general~\citep{jing2019self-supervised}. Indeed, computer vision models seem to benefit significantly from self-supervised learning as the amount of unlabeled data can be very large, while labeling can be expensive~\citep{jing2019self-supervised}. In this context, many visual pre-training tasks have been proposed, such as image rotation prediction~\citep{gidaris2018unsupervised}, colorization~\citep{colorization}, and solving jigsaw puzzles~\citep{jigsaw}. In addition to these context-based and generation-based pre-training methods, one can also leverage pseudo-labels from a pre-trained model in free semantic label-based methods~\citep{jing2019self-supervised}. In our work we therefore add a depth prediction pretext task, as advocated by~\citep{doersch2017multitask}, based on inferences from MegaDepth~\citep{li2018megadepth}. As the number of pretext tasks increases, so does the memory and computational time needed to process samples. This is especially problematic for generation-based methods which can be as computationally and memory intensive as the downstream task's model. We cannot therefore hope to train large models such as those used in I2IT, in parallel with all these tasks. \\

One must therefore derive a learning procedure which ensures that the networks do not forget as they change tasks: this is the focus of continual (or lifelong) learning. Neural networks have been plagued by the inability to maintain performance on previously accomplished tasks when they are trained on new ones -  a phenomenon that has been coined \textit{catastrophic forgetting}~\citep{kirkpatrick2017overcoming}. Various continual learning methods have been developed to mitigate forgetting which can be categorized as follows~\citep{lange2019continual}: replay-based methods, regularization-based methods and parameter isolation methods. In their work, \citet{matsumotocontinual} use the parameter-isolation method PiggyBack~\citep{mallya2018piggyback} in order to learn a sequence of I2IT tasks without forgetting the previous ones. \citet{zhai2019lifelong} on the other hand uses distillation \citep{hinton} in order to perform such tasks. In this work, we borrow ideas from the latter and apply them to a sequence of self-supervised tasks followed by a translation task.


\section{Approach}
\label{sec:approach}

\subsection{Model}
\label{subsec:model}
Our main contribution is a  continual learning framework which maintains self-supervised performance and improves unpaired I2IT representations. We chose as our I2IT model the simple and well-understood CycleGAN~\citep{cyclegan}.

Let  $\mathcal{T}$ be a set of $n = |\mathcal{T}|$ tasks such that $\{T^{(k)} | k < n-1 \}$ is a set of self-supervised tasks and $T^{(n-1)}$ is an I2IT task such as horse$\leftrightarrow$zebra~\citep{cyclegan}. The model is composed of two domain-specific sets of networks $M_A = \{G_A, D_A\}$ and $M_B = \{G_B, D_B \}$ where $G_i$ is a multi-headed generator and $D_i$ is a set of discriminators (1 per generative pretext task and 1 for the translation task). From now on, $i$ will be either $A$ or $B$ in which case $j$ is $B$ or $A$. All the following is symmetric in $i$ and $j$.

Let us focus on $G_i$. It is composed of an encoder $E_i$ and a set of $n$ task-specific heads $H_i^{(k)}$ which map from the encoder's output space to $T^{(k)}$'s output space. Let $x_i$ be a sample from domain $i$ and $z_i = E_i(x_i)$. In our work, we focus on the following 4 pretext tasks:

\begin{enumerate}
    \item $T^{(0)}$ is a rotation task, inspired from~\citep{gidaris2018unsupervised}, and $H_i^{(0)}$ performs a classification task: $H_i^{(0)}(z_i) \in [0, 1]^4$ as there are 4 possible rotations ($0^{\circ}$, $90^{\circ}$, $180^{\circ}$ and $270^{\circ}$). When appropriate (see \ref{subsec:schedule}) we train $H_i^{(0)}\circ E_i$ with a cross-entropy objective $\mathcal{L}^{(0)}$.

    \item $T^{(1)}$ is a jigsaw puzzle as introduced by~\citep{jigsaw}. We split the image into 9 equal sub-regions which we randomly reorder according to 64 pre-selected permutations (out of the $9!$ possible ones): $H_i^{(1)}(z_i) \in [0, 1]^{64}$. Similarly, we train $H_i^{(1)}\circ E_i$ with a cross-entropy objective $\mathcal{L}^{(1)}$.

    \item $T^{(2)}$ is a relative depth prediction task inspired from~\citep{jiang2017selfsupervised}: $H_i^{(2)}(z_i) \in \mathbb{R}^{h\times w}$. $H_i^{(2)}\circ E_i$ is trained with an $L_1$ objective $\mathcal{L}^{(2)}$ with respect to pseudo-labels obtained from a pre-trained MegaDepth model~\citep{li2018megadepth}.

    \item $T^{(3)}$ is a colorization task, as per~\citep{colorization, larsson2017colorproxy}: $H^{(3)}(z_i) \in [-1, 1]^{3 \times h \times w}$. Because a gray image can have several possible colorizations, we train $H_i^{(3)} \circ E_i$ by a mixture of $L^1$ loss with respect to $x_i$ and a GAN loss from a discriminator $D_i^{(3)}$: $\mathcal{L}^{(3)} = 0.1 L_1 + 0.9 L_{GAN}$
\end{enumerate}

The downstream translation task is based on CycleGAN's losses. For simplicity, in the following equations we call $G_i$ what is actually $H^{(4)}_i\circ E_i$, that is to say the standard CycleGAN generator.

\begin{align}
\begin{split}
L_{GAN} :{} & (G_i, D_i, x_i, x_j) \mapsto \mathbb{E}_{x_i}[\log(D_i(x_i))]\ +\\              &\mathbb{E}_{x_j}[\log(1 - D_i(G_i(x_j))))]
\end{split}\\
L_{idt}:{} & (G_i, x_i) \mapsto ||x_i - G_i(x_i) ||_1\\
L_{cyc}:{} & (G_i, G_j, x_i) \mapsto ||x_i - G_i(G_j(x_i)) ||_1\\
\mathcal{L}^{(4)} ={} & L_{GAN} + \lambda_{idt} L_{idt} + \lambda_{cyc}L_{cyc}
\end{align}

The overall model for domain $i$ is therefore composed of a shared encoder network and a set of $n$ heads which map from this latent space to their specific task's output space. We now need to understand how these tasks can be combined together in order to enable forward transfer from each of the self-supervised tasks to the translation task, without forgetting.

\subsection{Training Schedule}
\label{subsec:schedule}

When trying to incorporate self-supervised learning ideas into the I2IT framework, one could naively train all the heads in \textit{parallel} ($\lambda^{(k)}$ are scalars weighting the contribution of each loss):

\begin{equation}
    \mathcal{L}^{parallel} = \sum_{k=0}^{n-1} \lambda^{(k)} \mathcal{L}^{(k)}
\end{equation}

As explained previously, not only is this approach slower in that each sample has to go through all heads, but it also forces us to use smaller batch sizes for memory constraints.

Another naive approach would be to perform each task sequentially. Given an ordering $\tau$ of $\mathcal{T}$, one could train the model with:

\begin{equation}
    \mathcal{L}^{sequential}_{\tau(k)} = \lambda^{(\tau(k))} \mathcal{L}^{(\tau(k))}
\end{equation}

 In this \textit{sequential} training schedule, the model transitions from ($H_i^{(k)}, \mathcal{L}^{sequential}_{k}$) to ($H_i^{(k+1)}, \mathcal{L}^{sequential}_{k+1}$) according to some curriculum. For readability and without loss of generality, we omit $\tau$ from now on. In our experiments we implement a threshold-based curriculum where the transition from one task to the next depends on its performance (in both domains $A$ and $B$) on some validation metric (see Section \ref{sec:setup}).\\

In this paper we introduce Lifelong Self-Supervision, a \textit{continual} schedule which is similar to the sequential with the addition of a distillation loss. Inspired by \citet{tarvainen2017ean}, we maintain an exponential moving average of past encoder parameters, therefore keeping a weighted memory of all past encoders at the cost of single additional one. Formally, let $E_i^{(k)}$ be the frozen state of $E_i$ at the end of the $k^{th}$ task, \textit{i.e.} when transitioning from $T^{(k)}$ to $T^{(k+1)}$. Then we define the (non-trainable) reference encoder $\tilde{E_i}^{(k)}$ as follows:

\begin{equation}
    \label{eq:EMA}
      \tilde{E_i}^{(k)}= \Bigg\{\begin{array}{lr}
        0 & \text{if } k=0\\
         E_i^{(0)} & \text{if } k=1\\
        \alpha E_i^{(k-1)} + (1-\alpha) \tilde{E_i}^{(k-1)}  & \text{if } k>1
        \end{array}
\end{equation}

With $\alpha \in [0, 1]$. We use $\tilde{E_i}$ in an additional distillation term in the loss, minimizing the distance between the current and reference encoded spaces:

\begin{equation}
    L_{dist}^{(k)}: (\tilde{E_i}^{(k)}, E_i, x_i) \mapsto  ||\tilde{E_i}^{(k)}(x_i) - E_i(x_i)   ||_1
\end{equation}

\begin{equation}
    \mathcal{L}^{continual}_t =  \mathcal{L}^{sequential}_{t} + \beta L_{dist}^{(t)}
\end{equation}

These ideas are general and not specific to I2IT or CycleGAN ; this is why LiSS refers to $(\mathcal{T}, \mathcal{L}^{continual})$, not to a specific model.

\begin{figure}[!h]
    \centering
    \begin{minipage}[t]{0.45\textwidth}\centering%
        \includegraphics[width=\textwidth]{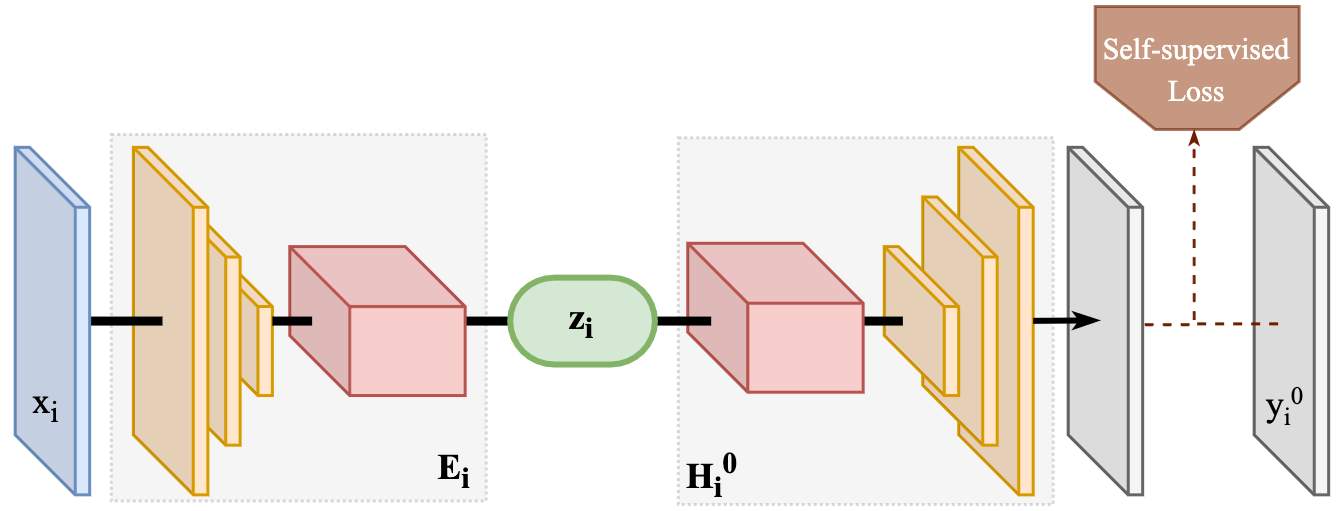}
        \caption{Training of the first pretext task $H_i^{(0)} \protect\circ E_i$. Orange blocks are (de-)convolutions, red blocks are sets of residual blocks. Here, $H_i^{(0)}$'s structure is for illustration purposes, see Appendix \ref{sec:appendix} for more details}
    \end{minipage}\hfill
    \begin{minipage}[t]{0.45\textwidth}\centering%
        \includegraphics[width=\textwidth]{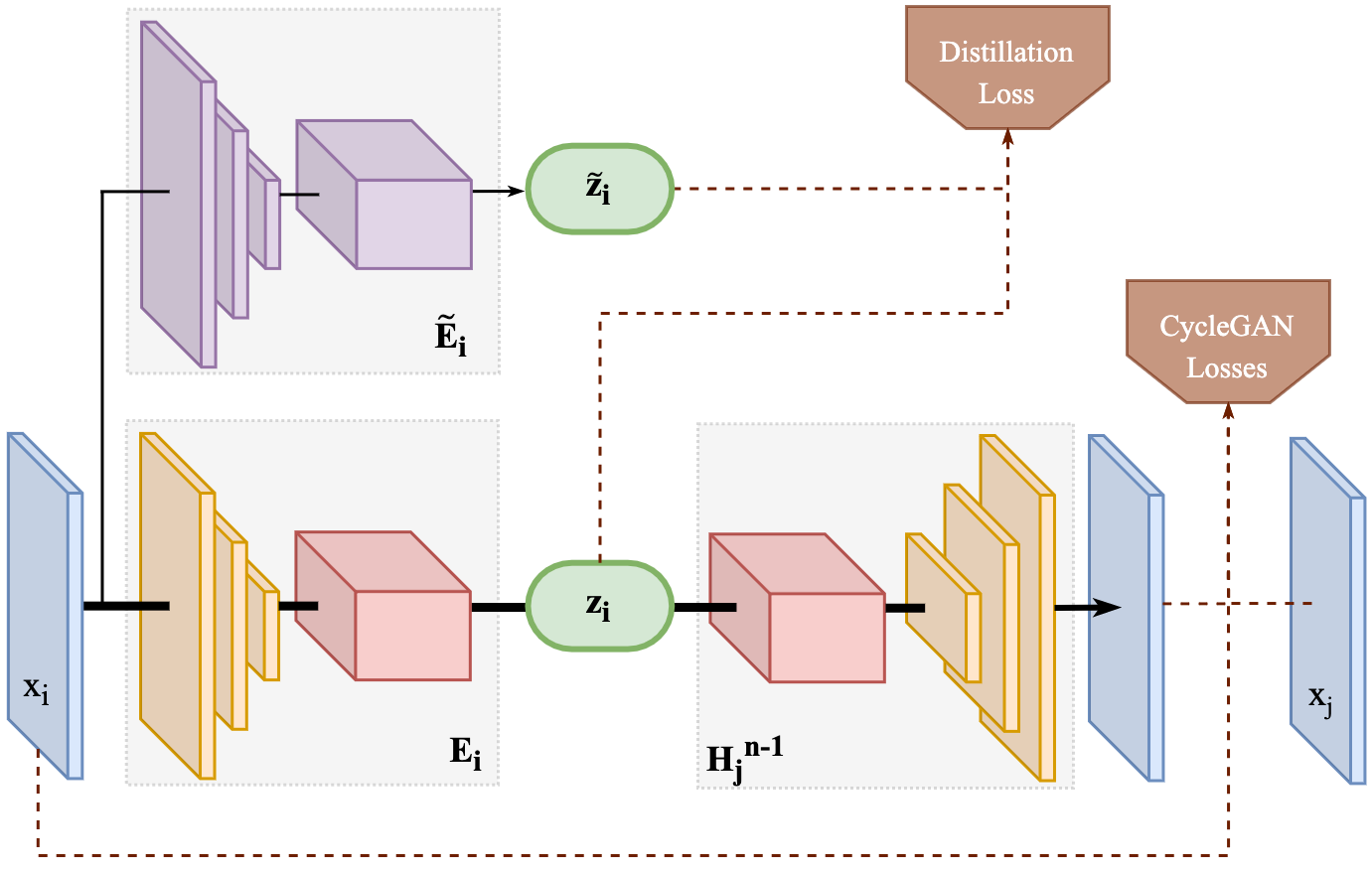}
        \caption{Training of the translation task $H_i^{(n-1)} \circ E_i$. As all tasks $T^{(k>0)}$, it includes a distillation loss between the current encoder's output $z$ and the reference encoder's, $z^{ref}$.}
    \end{minipage}
\end{figure}

\section{Experiments}
\label{sec:experiments}
\subsection{Setup}
\label{sec:setup}
To evaluate the effect of LiSS on CycleGAN's performance, we compare it with a baseline CycleGAN from~\citep{cyclegan} and to the two aforementioned naive training schedules: \textit{sequential} and \textit{parallel}. We compare these 4 models on the horse$\leftrightarrow$zebra dataset  on a dataset of flooded$\leftrightarrow$non-flooded street-level scenes from \citep{vicc} (the task is to simulate floods). As our goal is to understand how to efficiently leverage a set of given pretext tasks to improve representations, we keep $\mathcal{T}$ constant across experiments.\\

All models are trained with the same hyper-parameters. We use the RAdam optimizer~\citep{radam} with a learning rate of $0.0005$. We keep $\lambda_{idt}$ and $\lambda_{cyc}$ to their default values and set all $\lambda^{(k)}$ to $1$. We leave the analysis of $\alpha$'s exact impact for future work and set it to $0.5$ across experiments (see Eq.~\ref{eq:EMA}). Results are compared after $230$k translation steps. The continual and sequential models therefore have more total steps of training but in all cases $H^{(n-1)}_i$ is trained for $230$k steps (\textit{i.e.}  24 hours of training with the LiSS training schedule).
We set a fixed curriculum as per section \ref{subsec:model} with thresholds at 85\% for classification tasks, and $L_1$ distance of 0.15 for regression tasks. These were set to be $\sim$ 95\% of the parallel schedule's final validation performance. Batch-size is set to 5, for LiSS and \textit{sequential} schedules, but to 3 for the \textit{parallel} schedule\footnote{these are the largest values which fit in a Nvidia V100's 16GB of GPU memory.}.


\subsection{Image-to-Image Translation}
\begin{figure*}[!htbp]
\centering
\includegraphics[width=0.8\textwidth, height=9cm]{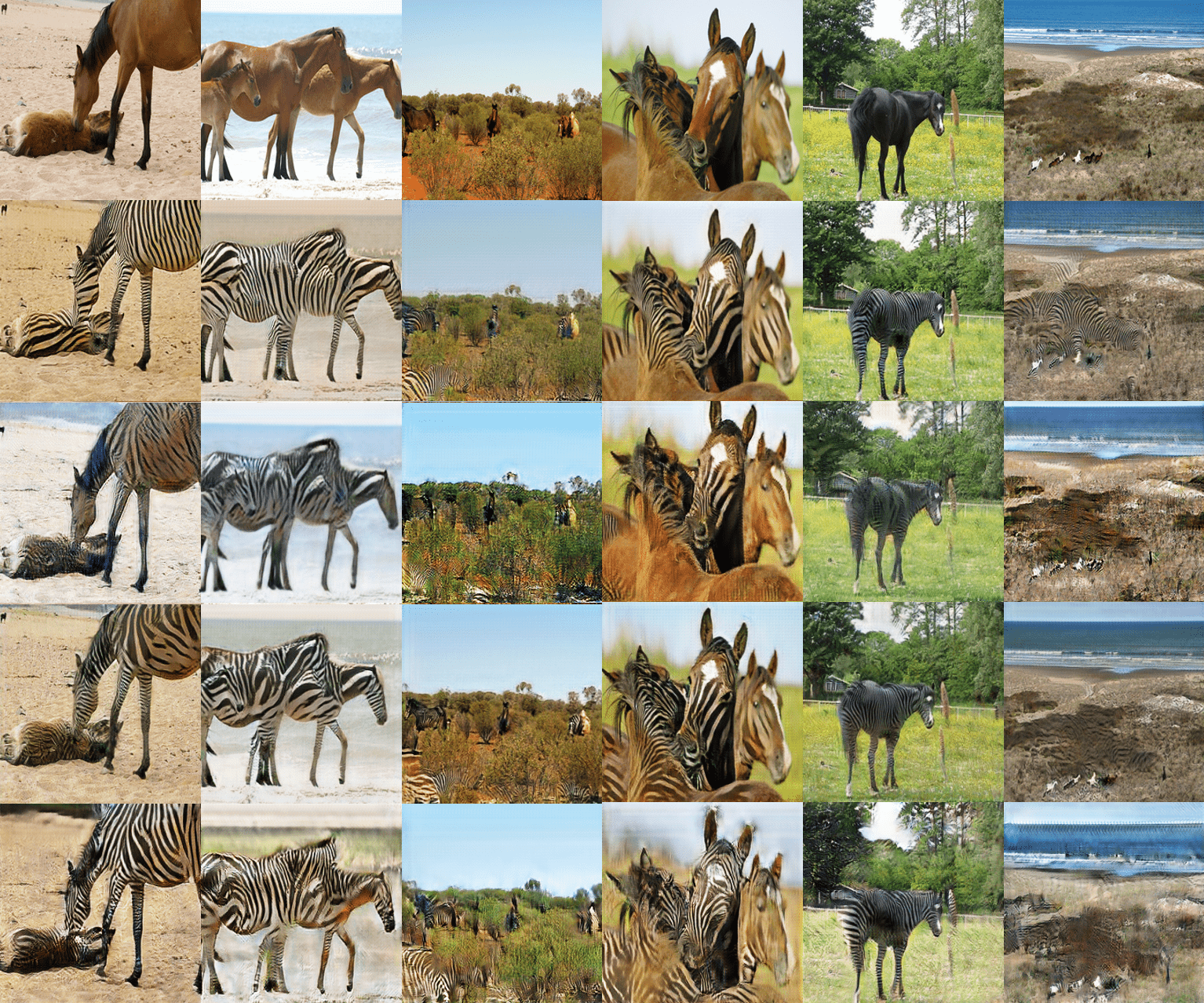}

   \caption{Comparison of models on the horse$\leftrightarrow$zebra dataset, with rows corresponding the image to translate (row 0) then translations from: CycleGAN (row 1), LiSS CycleGAN (row 2), Parallel Schedule (row 3), Sequential schedule(row 4). Note: the slight discrepancy in cropping is due to data-loading randomness}
\label{fig:h2z}
\vspace*{\floatsep}
\includegraphics[width=0.8\textwidth,height=9cm]{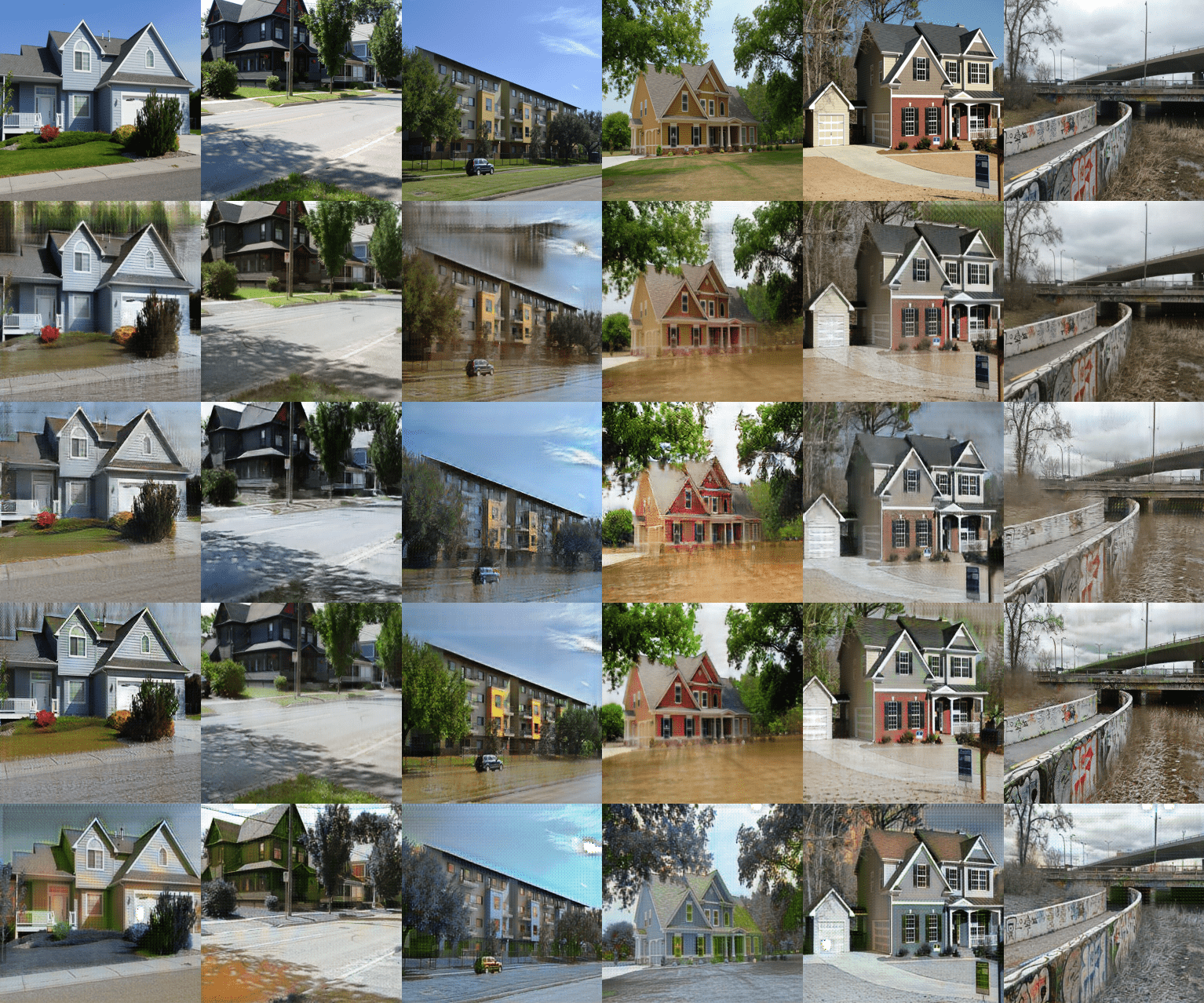}

   \caption{Comparison of models on the flooded$\leftrightarrow$non-flooded dataset, with rows corresponding the image to translate (row 0) then translations from: CycleGAN (row 1), LiSS training (row 2), parallel training (row 3), sequential training (row 4).}
\label{fig:floods}

\end{figure*}

Figure \ref{fig:h2z} and \ref{fig:floods} show how the LiSS framework visually fares against the other schedules. Images are referred to as $[i, j]$ meaning row $i$ and column $j$ in those figures.

While our setup does not quite match the pixelwise translation performance of CycleGAN, the model learns some interesting semantic features. Unlike CycleGAN which tends to merge distinct instances of entities that are very close to each other (Figure \ref{fig:h2z}'s image $[1, 1]$ for instance), our model is able to disentangle these instances and retain them as separate entities after translation. We can also see from Figure \ref{fig:h2z}'s image $[1, 4]$ and Figure \ref{fig:floods}'s images $[1, 0]$ and $[1, 2]$ that CycleGAN relies on color-based features, as evidenced by the zebra stripes projected on the brown patch of ground and the sky artificats. On the other hand, adding self-supervised tasks makes the models less sensitive to such glaring errors (see rows below the aforementioned CycleGAN translations in Figures \ref{fig:h2z} and \ref{fig:floods}).

Compared to the parallel schedule, LiSS keeps relevant features without enforcing a continuous training of $H_i^{(k<n-1)}$, which gives useful freedom to the model. It is able to have a similar translation performance and better color consistency, though one could argue that the parallel's translations are visually slightly better.

The sequential schedule on the other hand seems to have slightly worse translation performance. We can see that some of the useful "knowledge" the two other models still have is no longer available to the translation head as the smaller zebra is merged with the taller one in Figure \ref{fig:h2z}'s image $[4, 1]$ and the brown patch in image $[4, 4]$ shows slight stripes.

\subsection{Continual Learning Performance}

\begin{figure*}[h!]
    \centering
    \subfloat[Accuracies - LiSS]{
        \label{fig:liss_h2z_accuracies}
        \includegraphics[width=0.3\textwidth, height=0.19\textwidth]{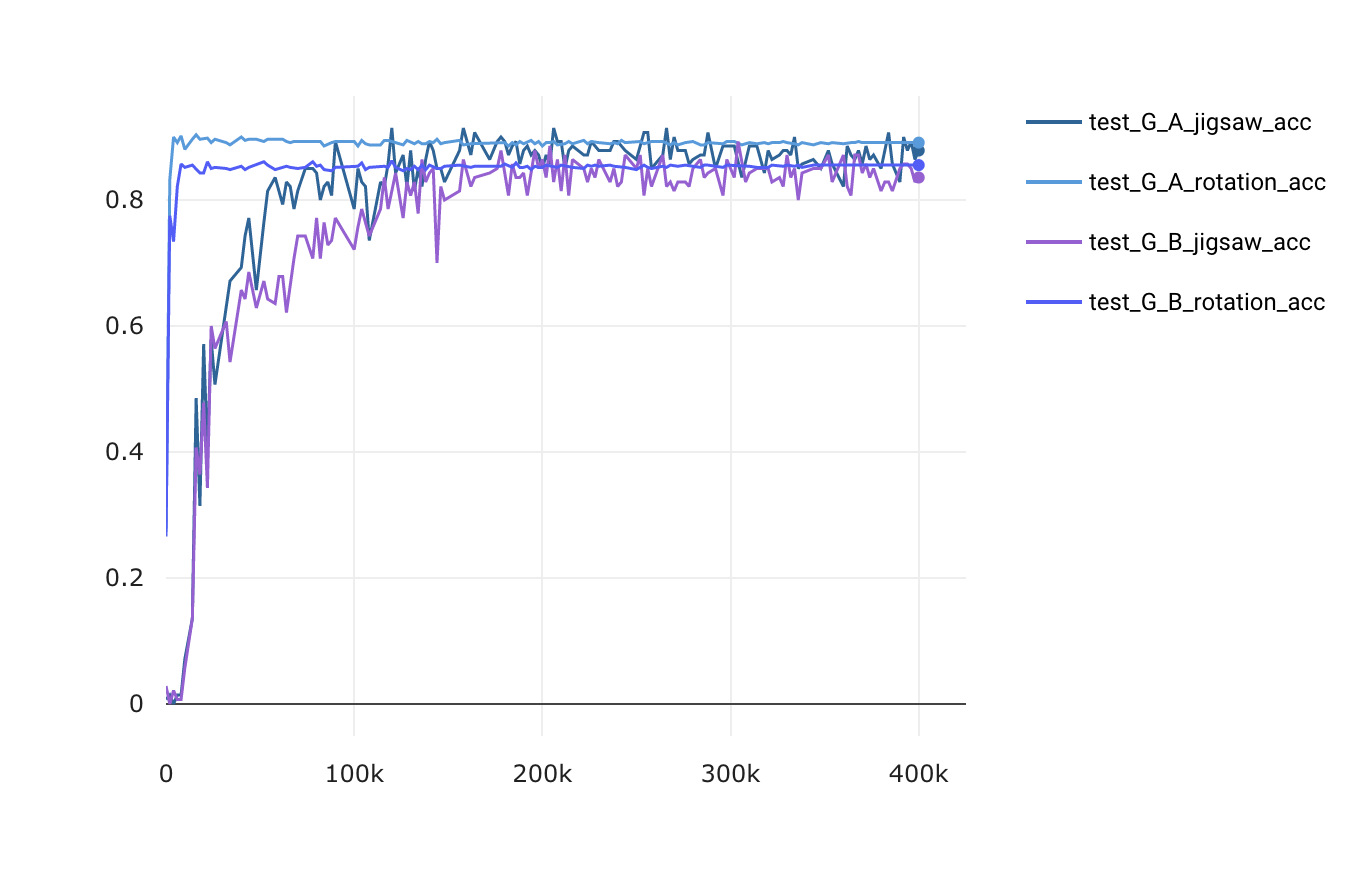}
    }
    \subfloat[Accuracies - Parallel]{
        \label{fig:parallel_h2z_accuracies}
        \includegraphics[width=0.3\textwidth, height=0.19\textwidth]{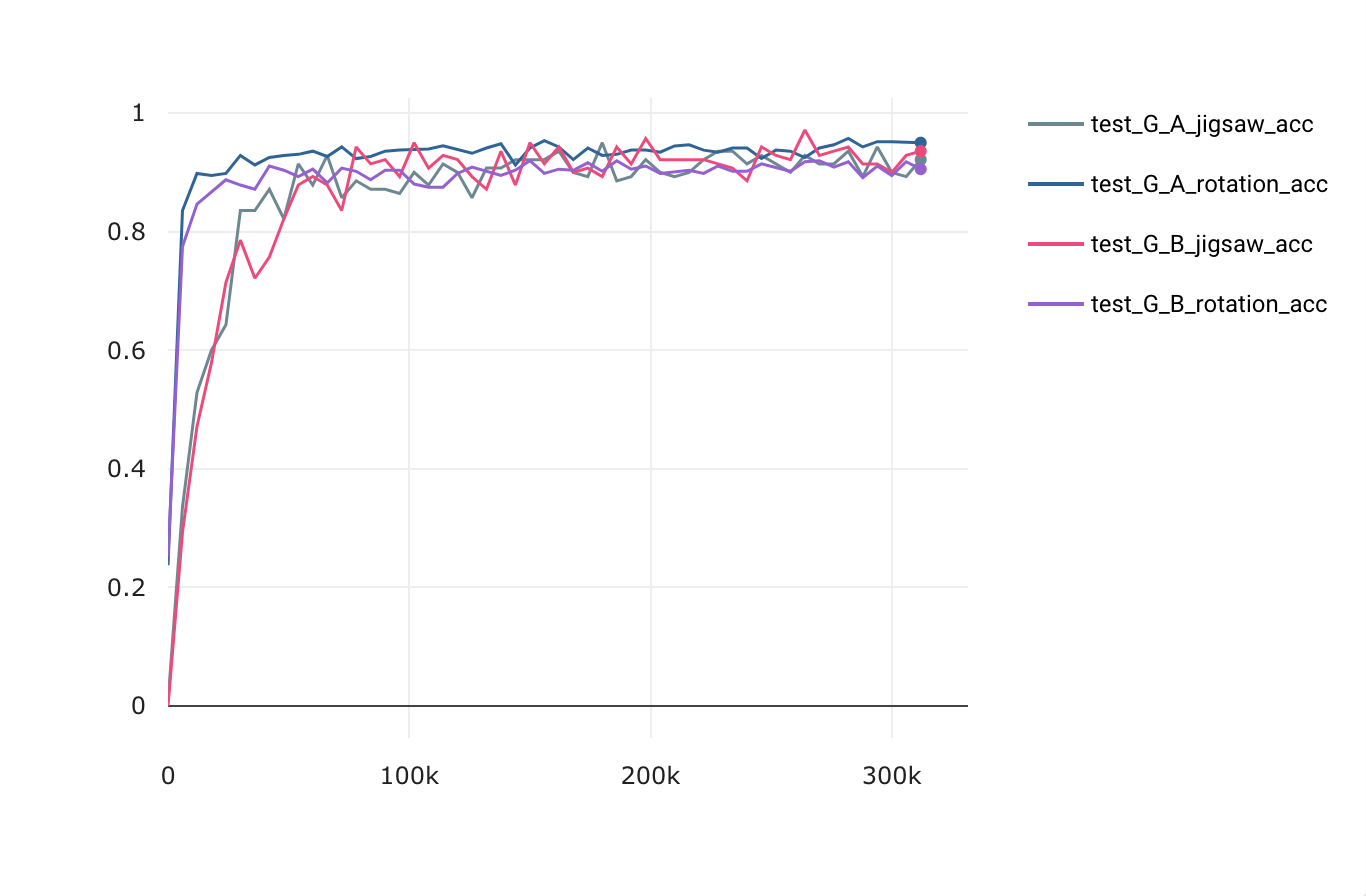}
    }
    \subfloat[Accuracies - Sequential]{
        \label{fig:sequential_h2z_accuracies}
        \includegraphics[width=0.3\textwidth, height=0.19\textwidth]{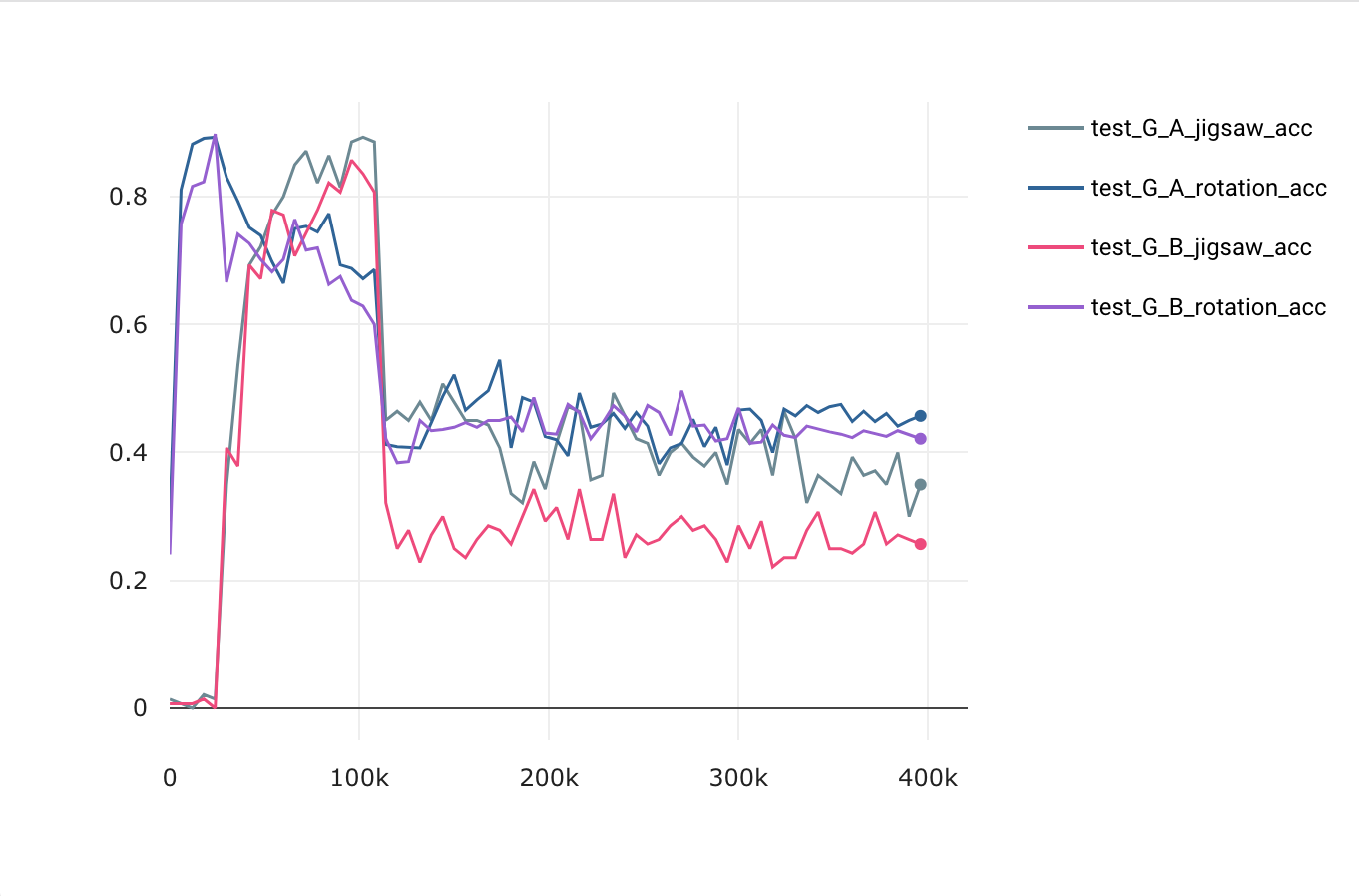}
    }
    \newline
    \subfloat[Losses - LiSS]{
        \label{fig:liss_h2z_loss}
        \includegraphics[width=0.3\textwidth, height=0.19\textwidth]{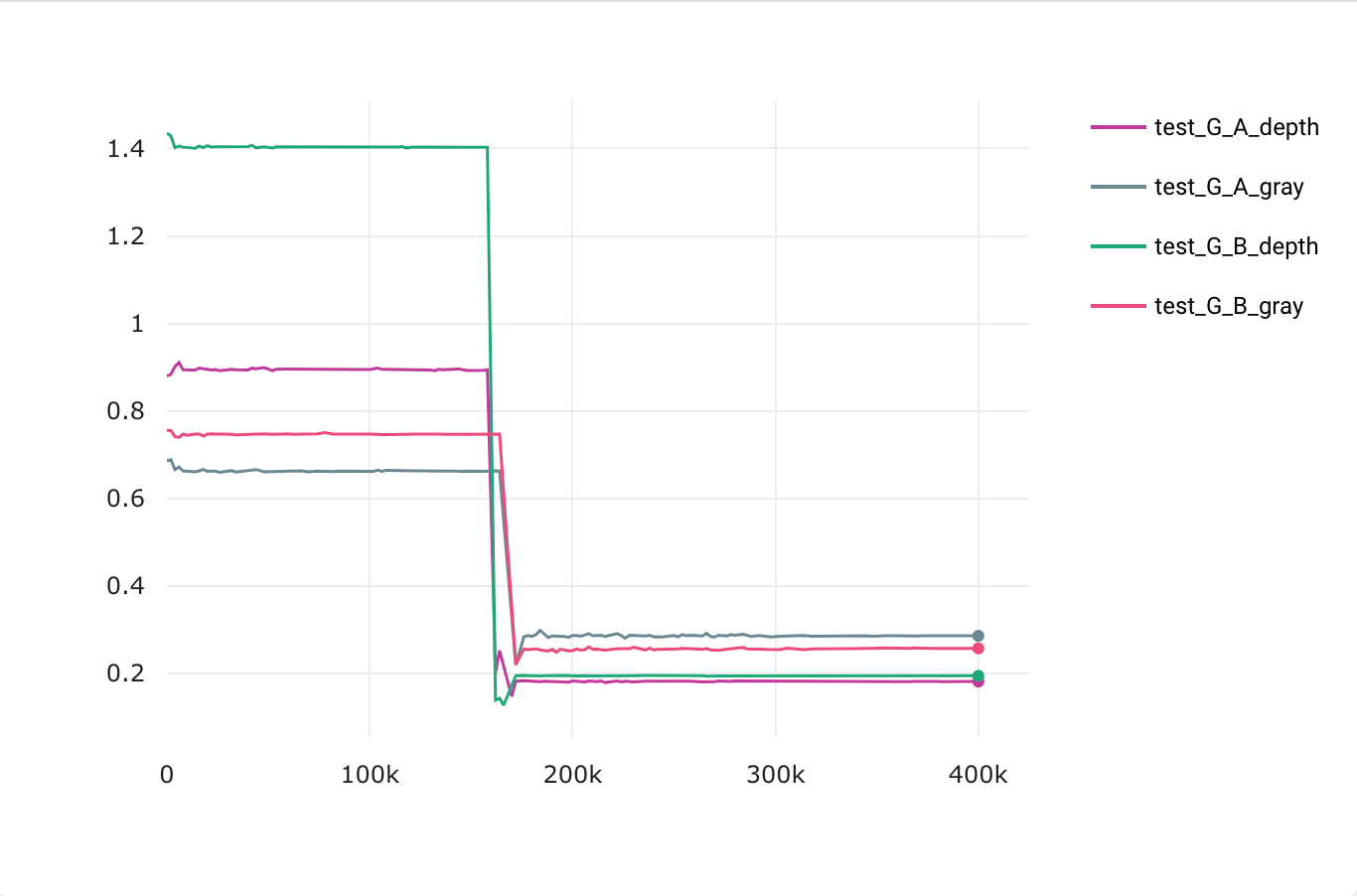}
    }
    \subfloat[Losses - Parallel]{
        \label{fig:parallel_h2z_loss}
        \includegraphics[width=0.3\textwidth, height=0.19\textwidth]{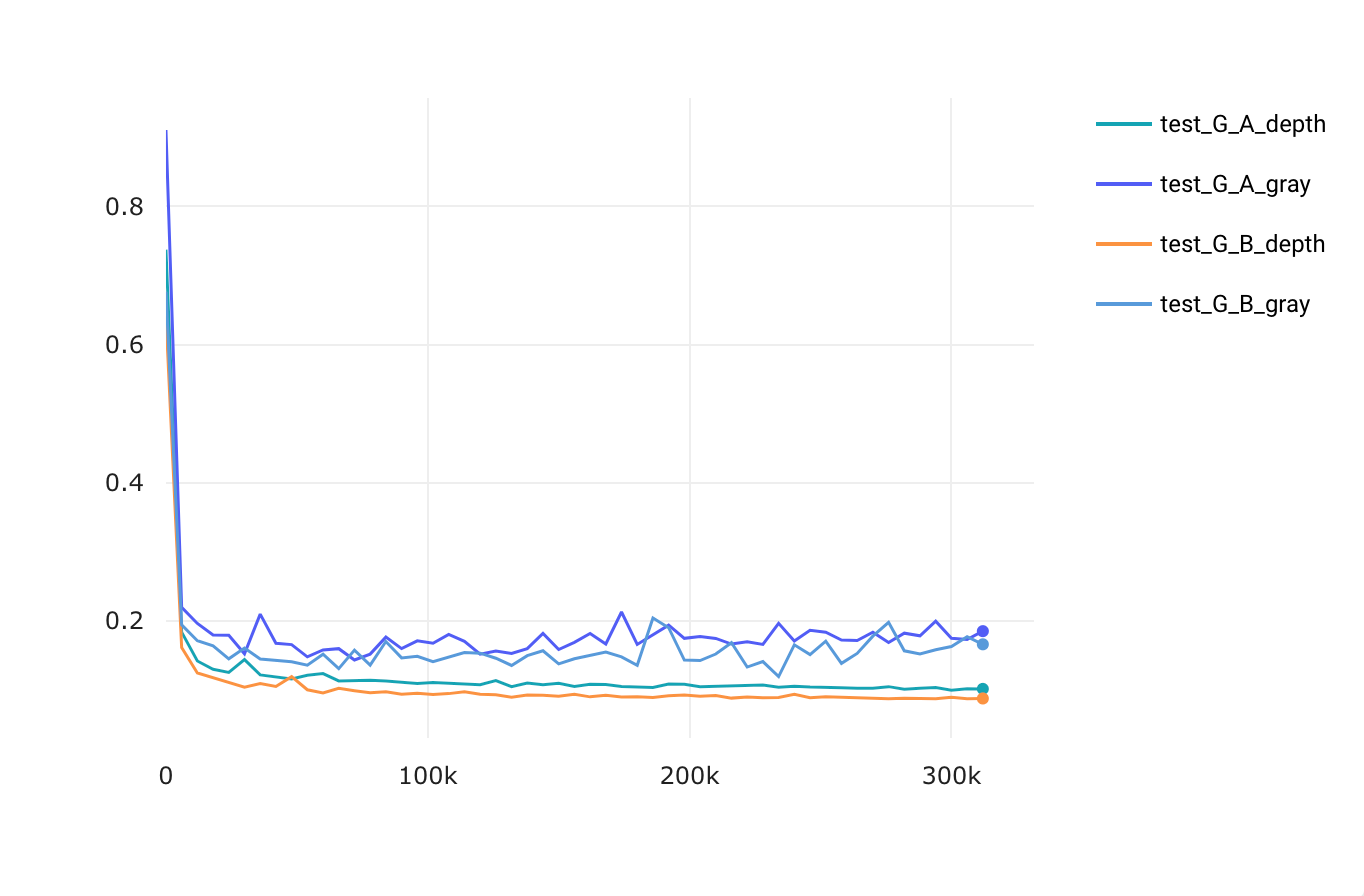}
    }
    \subfloat[Losses - Sequential]{
        \label{fig:sequential_h2z_loss}
        \includegraphics[width=0.3\textwidth, height=0.19\textwidth]{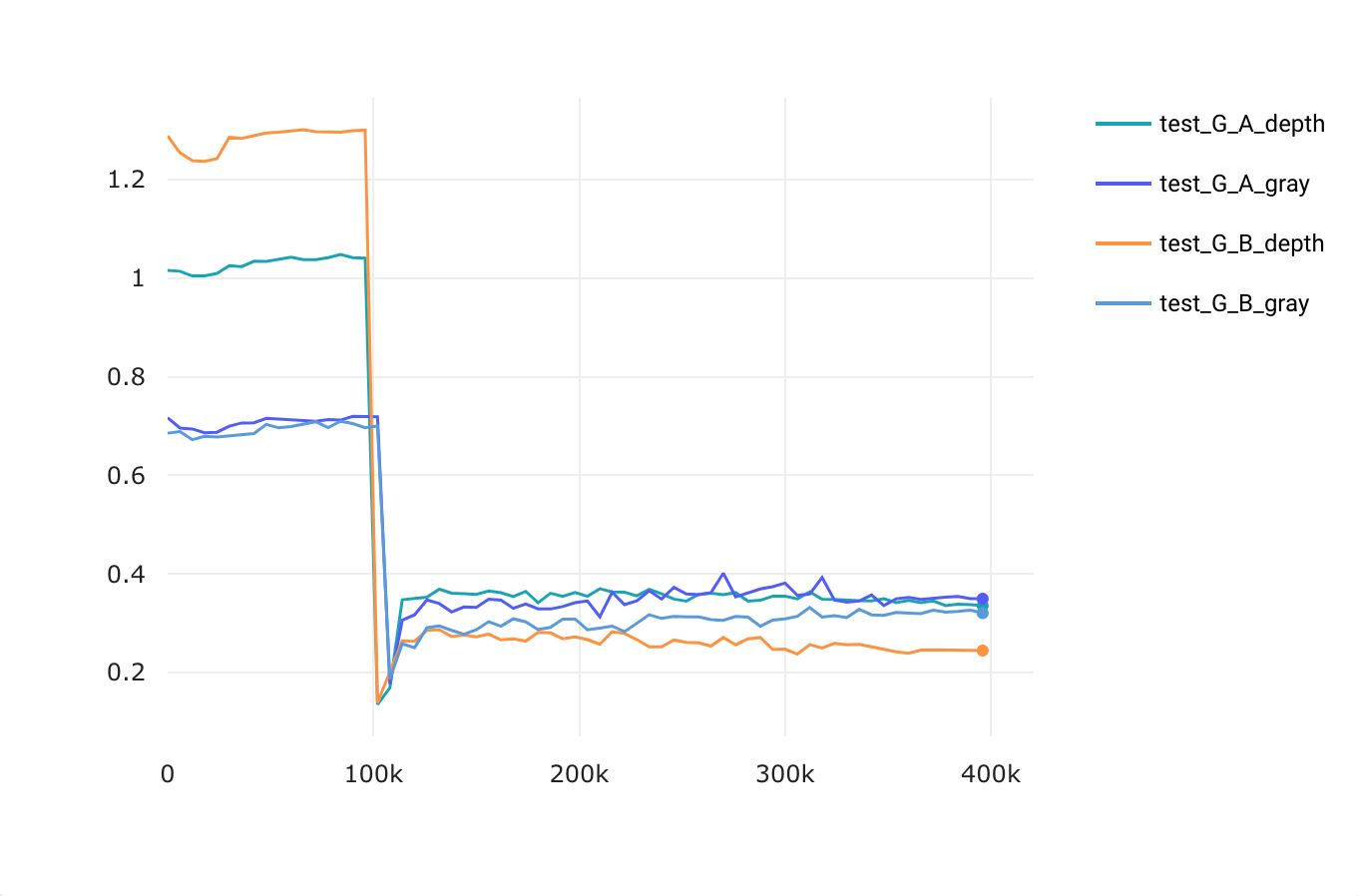}
    }
    \caption{
        Validation performance of the various schedules on the \textbf{horse$\protect\leftrightarrow$zebra} dataset. Accuracies are reported in the top row for the rotation and jigsaw heads of both $G_A$ and $G_B$. Similarly, colorization (named \textit{gray} in the plots) and depth prediction regression performances are plotted in the bottom row. Note how, unlike \textit{sequential} training, LiSS training maintains validation accuracies even though the model does not see the tasks anymore. Losses bump a little but converge to a better value than the sequential's. This illustrates how the LiSS training framework enables the model to leverage independent tasks' benefits while maintaining sufficient flexibility to learn new tasks, at a very low cost.
    }
     \subfloat[Accuracies - LiSS]{
        \label{fig:liss_floods_accuracies}
        \includegraphics[width=0.3\textwidth, height=0.19\textwidth]{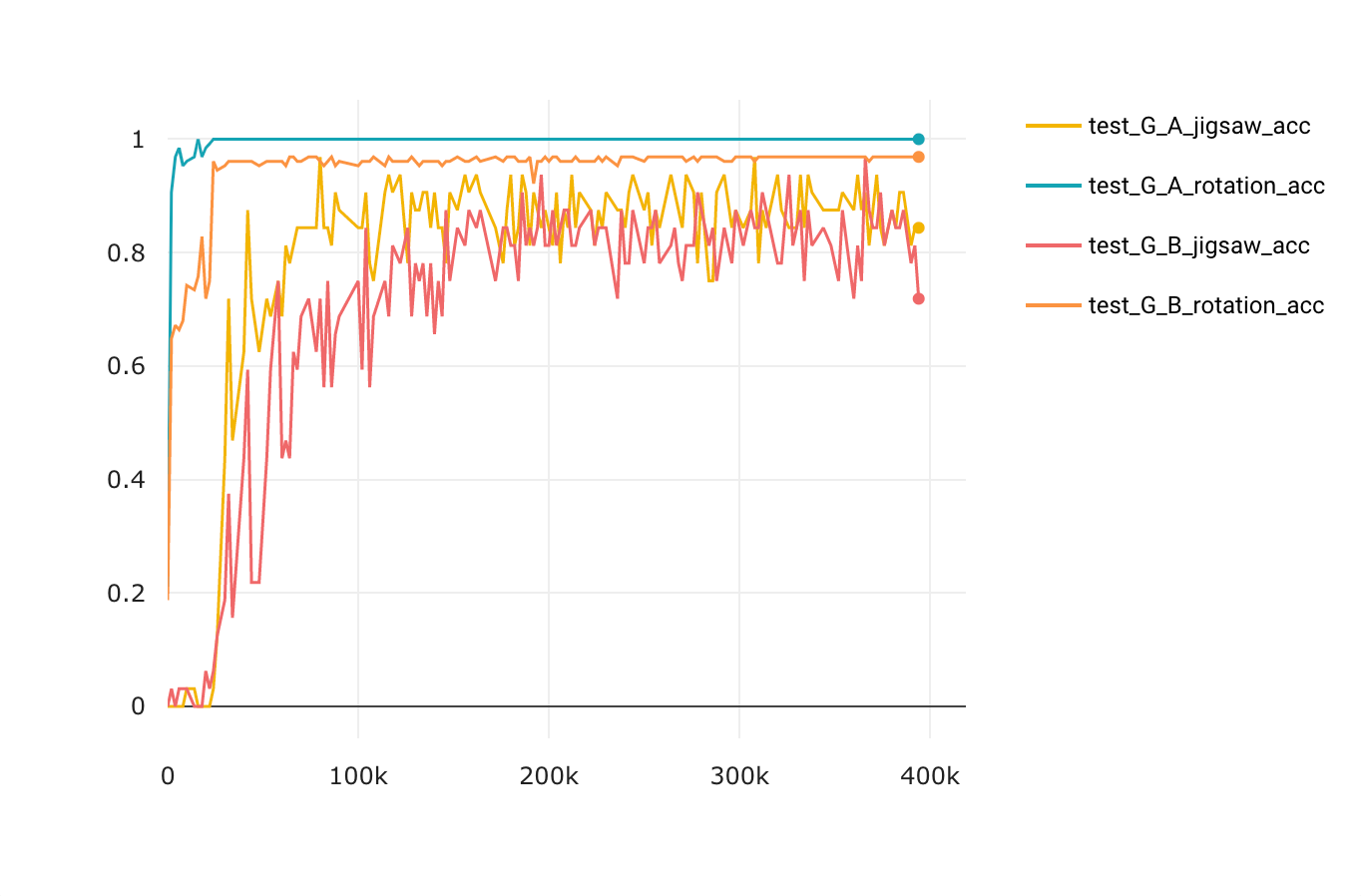}
    }
    \subfloat[Accuracies - Parallel]{
        \label{fig:parallel_floods_accuracies}
        \includegraphics[width=0.3\textwidth, height=0.19\textwidth]{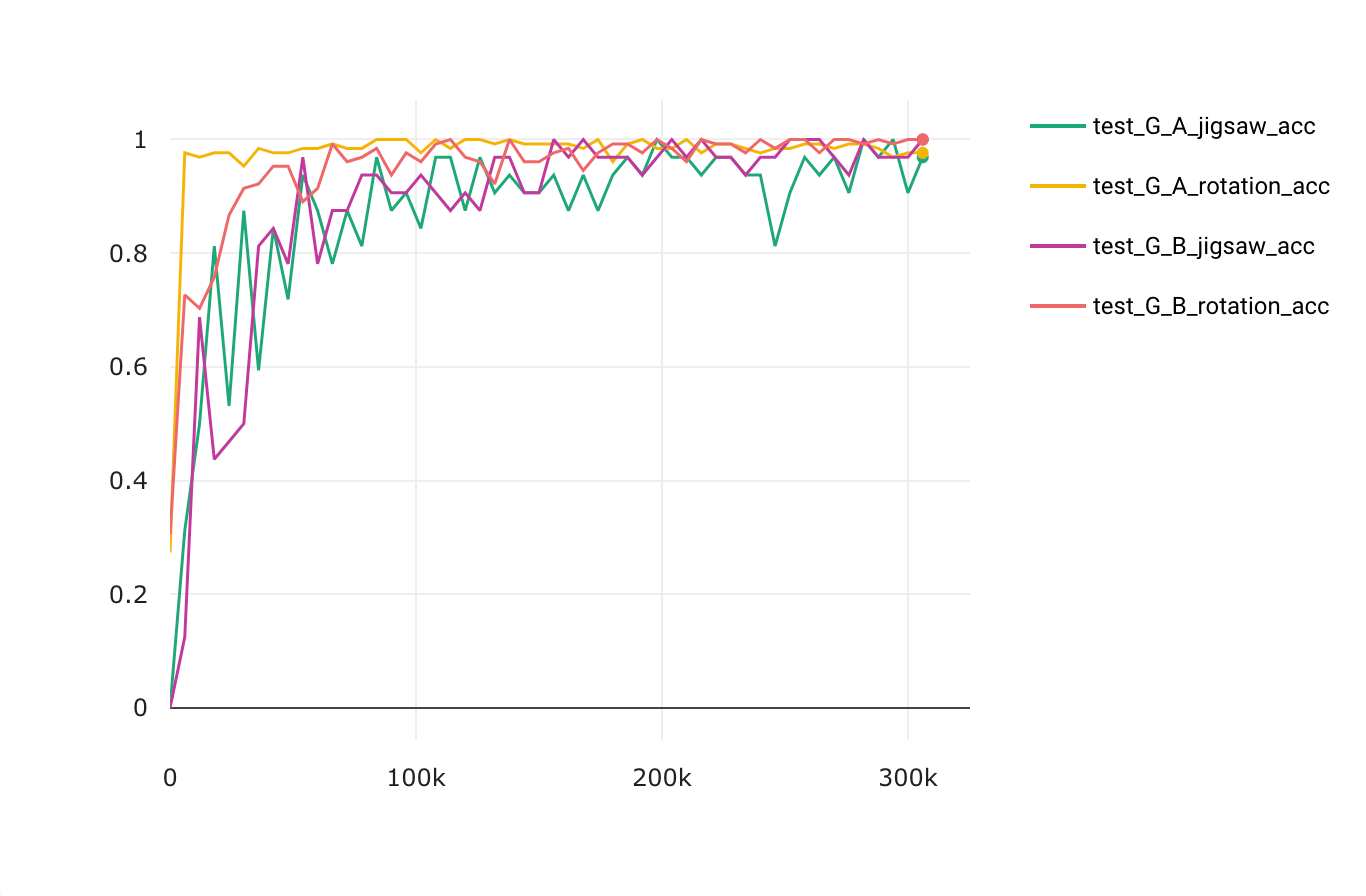}
    }
     \subfloat[Accuracies - Sequential]{
        \label{fig:sequential_floods_accuracies}
        \includegraphics[width=0.3\textwidth, height=0.19\textwidth]{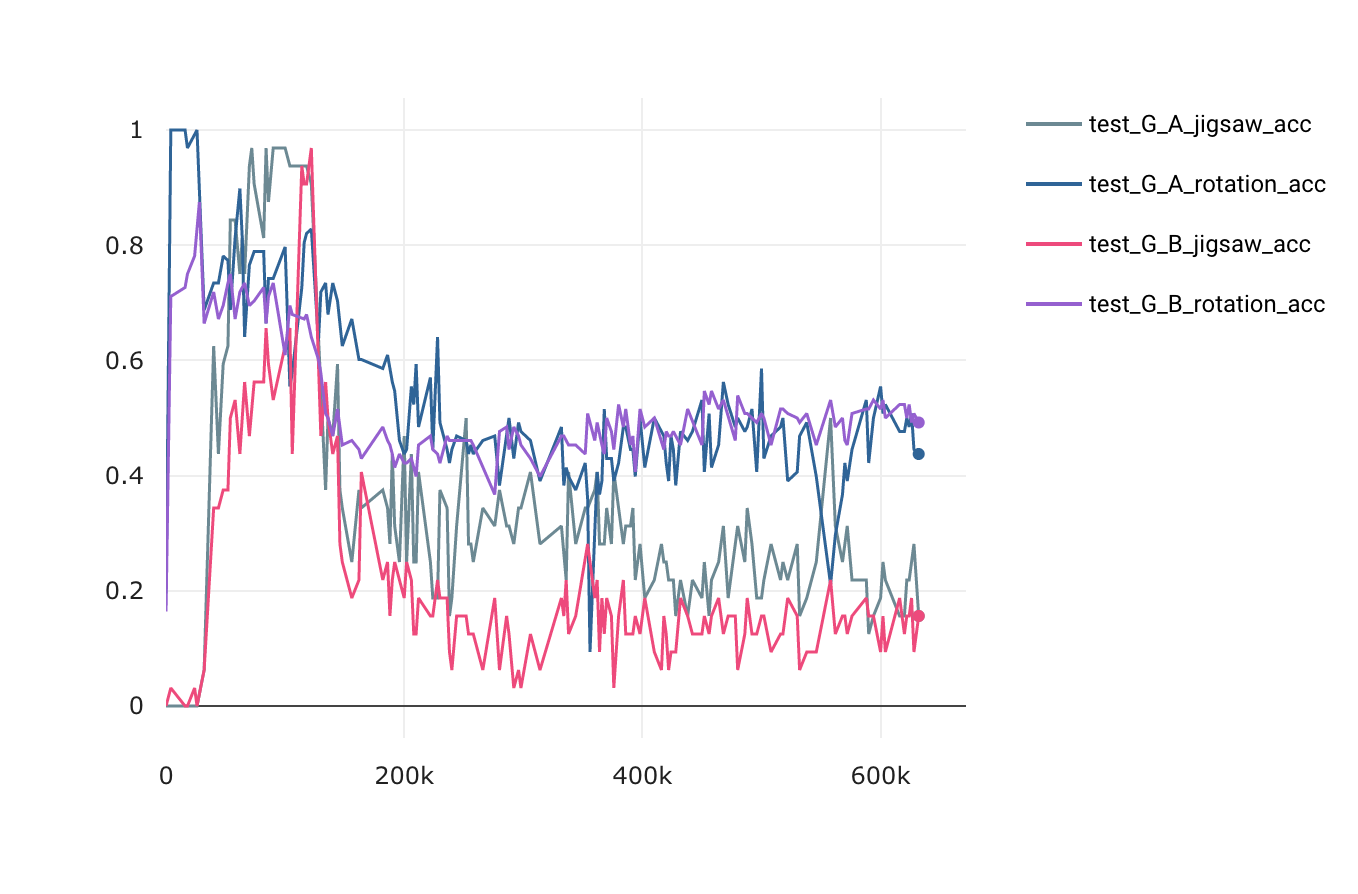}
    }
    \newline
     \subfloat[Losses - LiSS]{
        \label{fig:liss_floods_loss}
        \includegraphics[width=0.3\textwidth, height=0.19\textwidth]{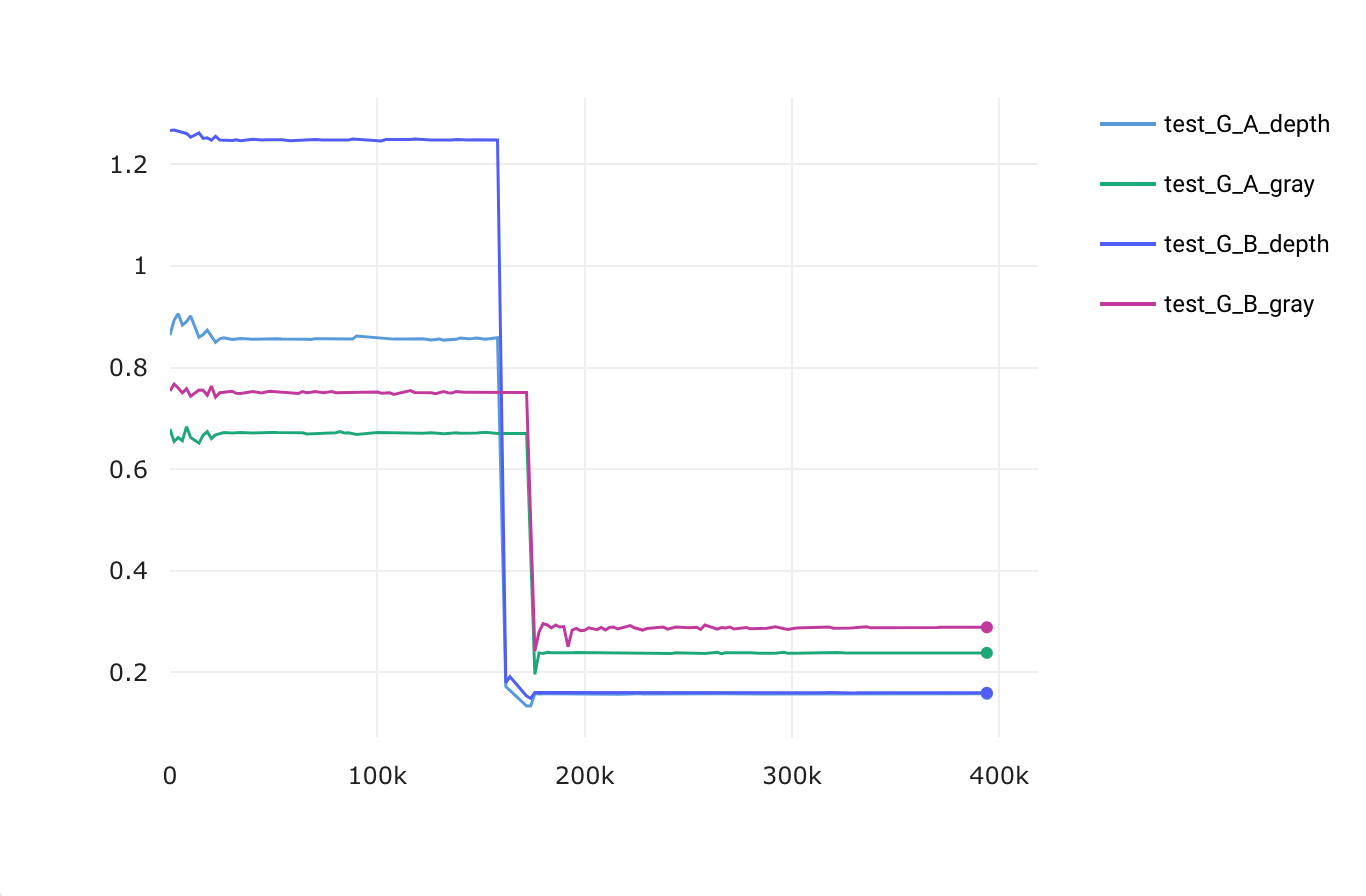}
    }
    \subfloat[Losses - Parallel]{
        \label{fig:parallel_floods_loss}
        \includegraphics[width=0.3\textwidth, height=0.19\textwidth]{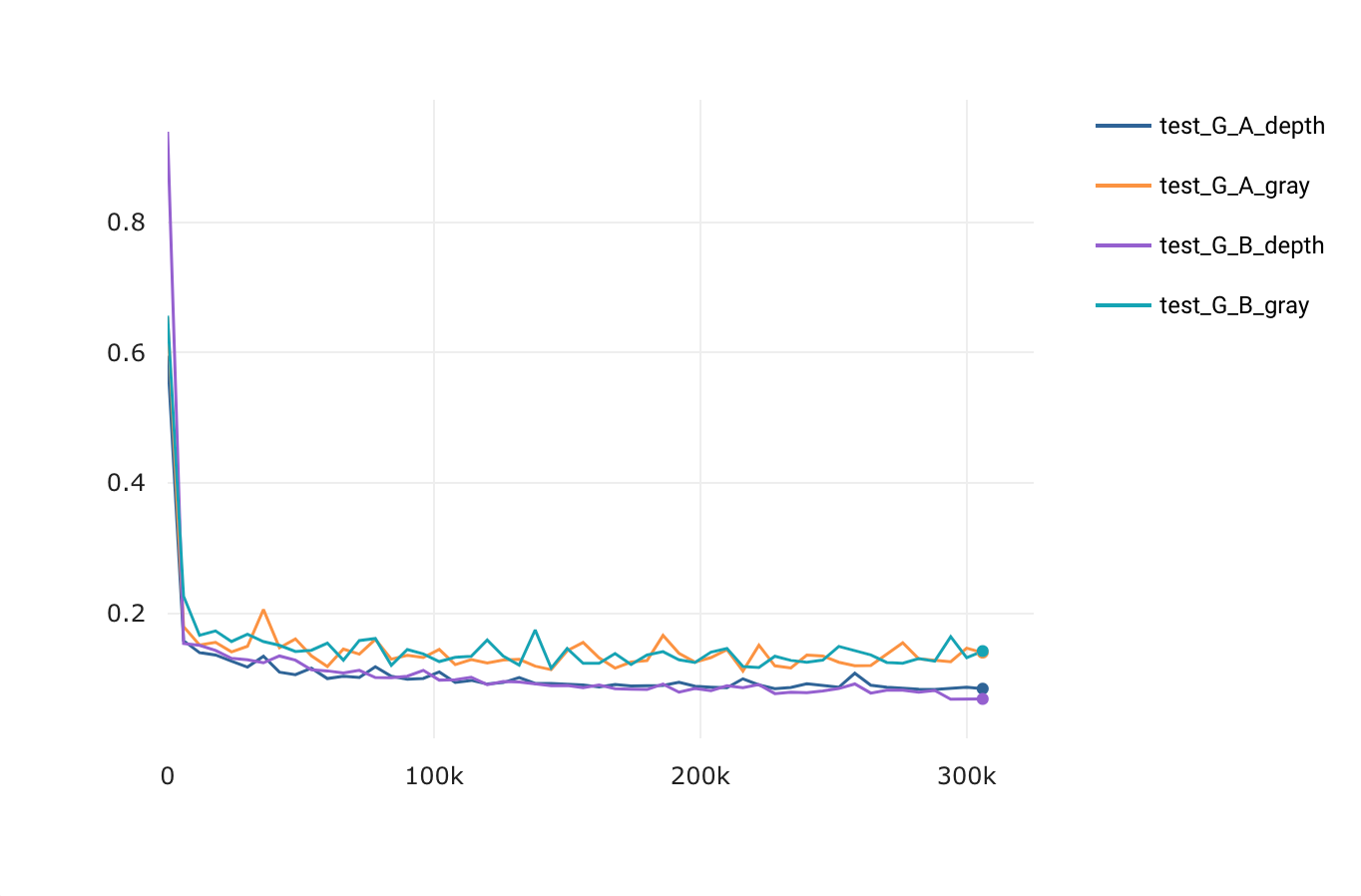}
    }
   \subfloat[Losses - Sequential]{
        \label{fig:sequential_floods_loss}
        \includegraphics[width=0.3\textwidth, height=0.19\textwidth]{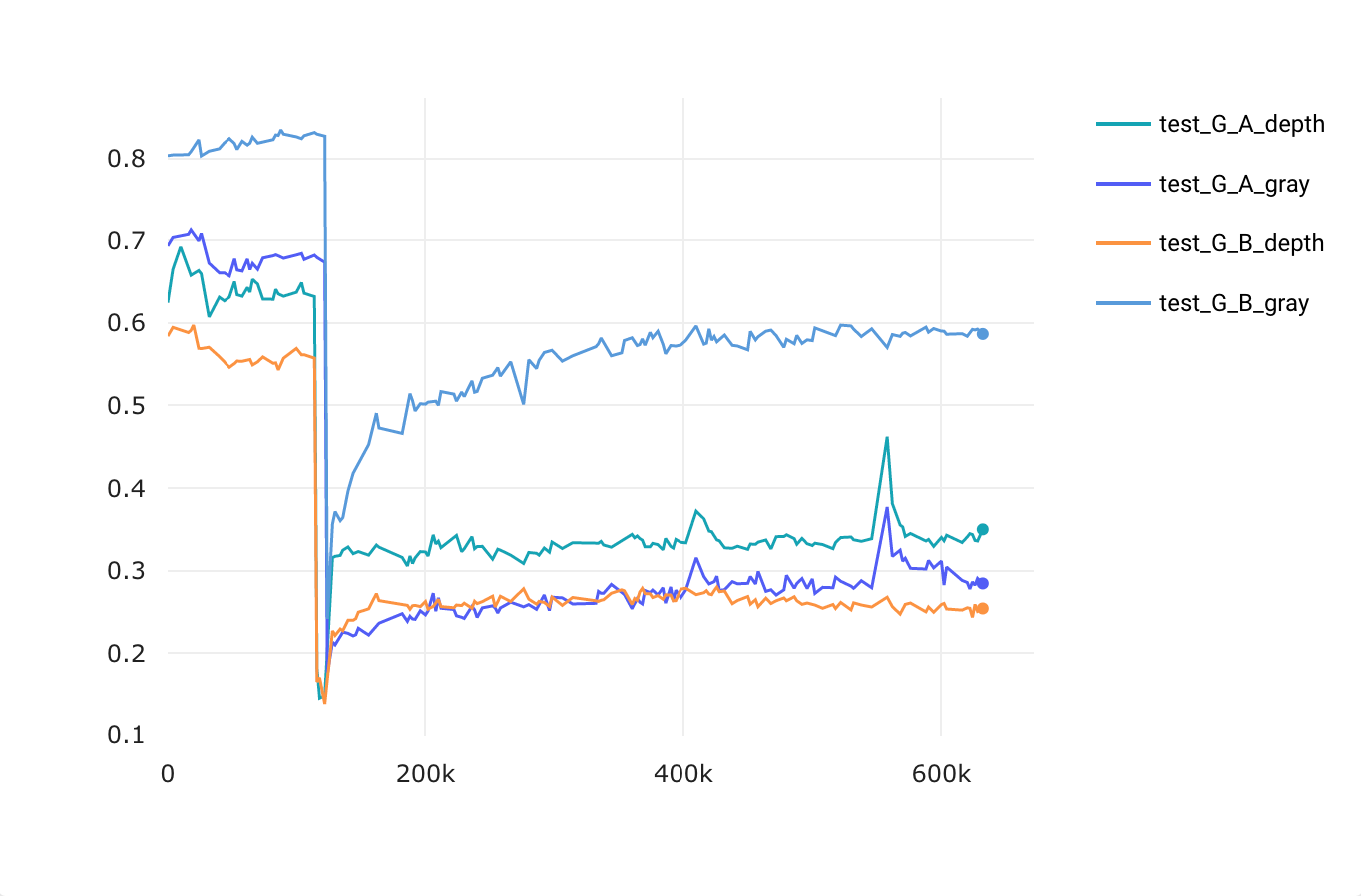}
    }
    \caption{
        Same plots as in Figure \ref{fig:h2zplots} for the \textbf{flooded$\protect\leftrightarrow$non-flooded} dataset. Once again we can see the drastic difference between LiSS and the naïve \textit{sequential} training schedule. The difference is much milder when comparing LiSS with \textit{parallel} training. The distillation loss prevents forgetting and maintains performance while allowing the network to learn new tasks. Transition steps are referenced in the Appendix's table \ref{table:floodstransitions}.
    }
    \label{fig:floodsplots}
    \label{fig:h2zplots}
\end{figure*}

Our main finding is that Lifelong Self-Supervision partially prevents forgetting.  We can see in figures 5 and \ref{fig:floodsplots} that our formulation preserves the model  from a forgetting as severe as in \textit{sequential} training while providing enough flexibility for it to learn new tasks. \\

In both datasets, we observe that the naive training schedules behave as expected: the \textit{sequential} one is able to learn new tasks the fastest as the model is less constrained. However, it is noticed that the sequential setup forgets previous tasks almost instantly as it changes its focus to a new task. On the other hand, the more constrained \textit{parallel} schedule shows that continuously training on tasks allows the model to master them all at once. This however comes at a memory and time cost as we could not fit more than 3 samples per batch (vs 5 for the other schedules), and the average processing time per sample is much larger (0.27s against an average of 0.12s for the other schedules). This means that to complete 230k translation steps, the \textit{parallel} schedule typically takes more than $17$h when LiSS only takes $12$h (counting all the pretext tasks).

Figures 5 and \ref{fig:floodsplots} show how LiSS maintains accuracies for the Rotation and Jigsaw tasks while performing slightly worse on the Depth prediction and Colorization tasks. As the encoder produces increasingly richer representations, the distillation loss prevents it from mapping input images to regions that would harm previous tasks. Because of our problem's sequential nature, decoding heads $H^{(k < n-1)}_i$ do not change after they have achieved the curriculum's required performance and the burden of producing stable yet informative features entirely relies on the encoders $E_i$ as the heads cannot adjust to its changes. \\




Table \ref{table:h2ztransitions} and \ref{table:floodstransitions} show that it takes more steps for the tasks to be learnt with LiSS. Intuitively, when training sequentially, the encoders are free to adjust exactly to the task. When training with LiSS, they are more constrained and it takes more iterations for them to reach the same performance on pretext tasks. This constraint is however pliable enough for encoders to adjust to new tasks.

\begin{table}[!htb]
\centering
\begin{tabular}{@{}llll@{}}
\toprule
Schedule   & Task         & Start\_Step  & End\_Step \\ \midrule
LiSS       & Rotation     & 0            & 8 000      \\
           & Jigsaw       & 8 000        & 158 000    \\
           & Depth        & 158 000      & 170 000    \\
           & Colorization & 170 000      & 172 000    \\
\midrule
Sequential & Rotation     & 0            & 24 000     \\
           & Jigsaw       & 24 000       & 96 000     \\
           & Depth        & 96 000       & 102 000    \\
           & Colorization & 102 000      & 108 000    \\ \bottomrule
\end{tabular}
\caption{Transition steps for the horse$\protect\leftrightarrow$zebra task. Translation starts when the colorization task is mastered.}
\label{table:h2ztransitions}
\end{table}

\section{Discussion}
\label{sec:conclusion}
We propose a method, Lifelong Self-Supervision (LiSS), enabling CycleGAN to leverage sequential self-supervised auxiliary tasks to improve its representations. By distilling the knowledge of a reference encoder (which is an exponential moving average of previous encoders, in parameter space) we prevent catastrophic forgetting of the auxiliary tasks, thereby allowing CycleGAN to better disentangle instances of objects to be translated and rely less on colors. This framework can bring benefits of training on all the tasks at once at a much lower memory and computational cost as it only requires us to keep one additional encoder. Our exploratory experiments show encouraging results which will need further investigation in future work to produce a principled framework.

Open questions include the exact impact of the reference encoder's algebra (namely exponential moving average versus other moving averages and the impact of $\alpha$), a more thorough hyper-parameter search in order to tune $\lambda^{(k)}$ and $\beta$ and achieve better pixel-level results. Additionally, exploring schedules and auxiliary tasks would allow for a better understanding of how SSL can improve unpaired I2IT models. Finally, while CycleGAN's simplicity allowed us to isolate LiSS's contribution to improved translations, exploring its capabilities on more complex architecture is a promising direction for future work.


\newpage
{\small
\setlength{\bibsep}{0pt}
\bibliographystyle{abbrvnat}
\bibliography{_continual}
}
\appendix
\section{Implementation details}
\label{sec:appendix}



\begin{table}[]
\begin{tabular}{llll}
\toprule
Model      & Task         & Start\_Step  & End\_Step \\ \midrule
LiSS       & Rotation     & 0            & 24 000     \\
           & Jigsaw       & 24 000       & 158 000    \\
           & Depth        & 158 000      & 174 000    \\
           & Colorization & 174 000      & 176 000    \\
\midrule
Sequential & Rotation     & 0            & 28 000     \\
           & Jigsaw       & 28 000       & 114 000    \\
           & Depth        & 114 000      & 122 000    \\
           & Colorization & 122 000      & 124 000    \\ \bottomrule
\end{tabular}
\caption{Transition steps for the flooded$\leftrightarrow$non-flooded task. Translation starts when the colorization task is mastered.}
\label{table:floodstransitions}
\end{table}

Our framework's network architecture follows the baseline CycleGAN~\citep{cyclegan} with some differences in the generator to support self supervision.
We use ``ResnetBlock'' to denote residual blocks~\citep{residual_blocks}.
``C$\times$H$\times$W-S-P Conv'' represents a convolutional layer with C channels having kernel size H$\times$W with padding P and stride S.
``NConv'' denotes a convolutional layer followed by an instance norm.
``TConv'' denotes transpose convolution layer proposed by~\citet{conv_transpose} followed by instance norm.

\paragraph{Discriminator Network Architecture.}

We use $70\times70$ PatchGANs~\citep{isola2017image_patchgan, li2016precomputed_patchgan, ledig2017photo_patchgan} as the one used in the original CycleGAN~\cite{cyclegan} baseline model shown in Table \ref{table:discriminatorarch}. The discriminator's output is a real or fake label for overlapping $70\times70$ patches. The GAN loss function then compares the target's label real or fake to the average of patches predictions of the input image.
\begin{table}[ht]
\caption{Discriminator's PatchGAN Architecture}
\label{table:discriminatorarch}
\resizebox{\columnwidth}{!}{
\begin{tabular}{@{}lll@{}}
\toprule
Layer                                           & Output           & Activation\\ \midrule
Input                                           & $3\times256\times256$  & None               \\
$64\times4\times4-2-1$ Conv  & $64\times128\times128$ & LeakyReLU\\
$128\times4\times4-2-1$ NConv & $128\times64\times64$ & LeakyReLU \\
$256\times4\times4-2-1$ NConv  & $256\times32\times32$ & LeakyReLU  \\
$512\times4\times4-2-1$ NConv & $512\times31\times31$  & LeakyReLU \\
$1\times4\times4-2-1$ Conv   & $1\times30\times30$ & None    \\ \bottomrule
\end{tabular}
}
\end{table}

\paragraph{Encoder Network Architecture.}
The encoder network's architecture is inspired from~\citep{johnson2016perceptual}, as shown in Table \ref{table:encoderarch}. The network starts with a reflection padding of size 3 and zero padded 7x7 convolutions to avoid severe artifacts around the borders of the generated images, followed by 3x3 convolutional blocks  with padding 1 and stride 2 to downsample the input image and finally by 3 Residual Blocks.
\begin{table}[ht]
\caption{Encoder's Network Architecture}
\label{table:encoderarch}
\resizebox{\columnwidth}{!}{
\begin{tabular}{@{}lll@{}}
\toprule
Layer                                           & Output           & Activation\\ \midrule
Input                                           & $3\times256\times256$  & None               \\
ReflectionPad p=3  & $3\times262\times262$ & None\\
$64\times7\times7-1-0$ NConv & $64\times256\times256$ & ReLU \\
$128\times3\times3-2-1$ NConv  & $128\times128\times128$ & ReLU  \\
$256\times3\times3-2-1$ NConv  & $256\times64\times64$ & ReLU  \\
ResnetBlock  & $256\times64\times64$ & None  \\
ResnetBlock  & $256\times64\times64$ & None  \\
ResnetBlock  & $256\times64\times64$ & None  \\ \bottomrule
\end{tabular}
}
\end{table}

\paragraph{Translation and Colorization Head Architectures}
The translation head's network's architecture follows the standard CycleGAN generator~\citep{cyclegan} as shown in Table \ref{table:decoderarch}.
It consists of 3 residual blocks followed by upsampling convolutions. For colorization to share the encoder with other tasks, we repeat gray scale images along the channel dimension.

\begin{table}[ht]
\caption{Decoder and Colorization Network Architecture }
\label{table:decoderarch}
\resizebox{\columnwidth}{!}{
\begin{tabular}{@{}lll@{}}
\toprule
Layer                                           & Output           & Activation\\ \midrule
Input                                           & $256\times64\times64$  & None               \\
ResnetBlock  & $256\times64\times64$ & None  \\
ResnetBlock  & $256\times64\times64$ & None  \\
ResnetBlock  & $256\times64\times64$ & None  \\
$128\times3\times3-2-1$ TConv  & $256\times128\times128$ & ReLU  \\
$64\times3\times3-2-1$ TConv  & $64\times256\times256$ & ReLU  \\
ReflectionPad p=3 & $64\times262\times262$ & None \\
$3\times7\times7-1-0$ Conv & $3\times256\times246$ & Tanh \\ \bottomrule
\end{tabular}
}
\end{table}

\paragraph{Rotation Network Architecture.}
The rotation head's architecture is inspired from~\citep{gidaris2018unsupervised} and shown in Table \ref{table:rotarch}.
The network performs a simple classification task out of 4 possible rotations ($0^{\circ}$, $90^{\circ}$, $180^{\circ}$ and $270^{\circ}$).

\begin{table}[ht]
\caption{Rotation Network Architecture}
\label{table:rotarch}
\resizebox{\columnwidth}{!}{
\begin{tabular}{@{}lll@{}}
\toprule
Layer                                           & Output           & Activation\\ \midrule
Input                                           & $256\times64\times64$  & None               \\
$256\times3\times3-2-1$ NConv & $3\times32\times32$ & LeakyReLU \\
$256\times3\times3-2-1$ NConv & $3\times16\times16$ & LeakyReLU \\
$2\times2$ MaxPool & $3\times8\time8$ & None \\
$128\times3\times3-2-1$ NConv & $3\times4\times4$ & LeakyReLU \\
Flatten & $2048$ & None \\
$2048\times4$ Linear & $4$ & None \\ \bottomrule
\end{tabular}
}
\end{table}

\paragraph{Jigsaw Network Architecture.}
Jigsaw's network predicts the correct indices order of shuffled patches of an input image.
The network consists of a set of convolutions extracting useful features from input image and then a fully connected layer to map it to the possible permutations.
The model's architecture shown in Table \ref{table:jigsawarch} performs a classification task over 64 possible  permutations of shuffled images order.
\begin{table}[ht]
\caption{Jigsaw Network Architecture}
\label{table:jigsawarch}
\resizebox{\columnwidth}{!}{
\begin{tabular}{@{}lll@{}}
\toprule
Layer                                           & Output           & Activation\\ \midrule
Input                                           & $256\times64\times64$  & None               \\
$256\times3\times3-2-1$ NConv & $3\times32\times32$ & LeakyReLU \\
$256\times3\times3-2-1$ NConv & $3\times16\times16$ & LeakyReLU \\
$2\times2$ MaxPool & $3\times8\time8$ & None \\
$128\times3\times3-2-1$ NConv & $3\times4\times4$ & LeakyReLU \\
Flatten & $2048$ & None \\
$2048\times64$ Linear & $64$ & None \\ \bottomrule
\end{tabular}
}
\end{table}

\paragraph{Depth Prediction Network Architecture.}
Depth network architecture is inspired from~\citep{jiang2017selfsupervised} and shown in Table \ref{table:deptharch}.
The network is trained on labels predicted using a pre-trained MegaDepth Model~\citep{li2018megadepth}.

\begin{table}[ht]
\caption{Depth Prediction's Network Architecture}
\label{table:deptharch}
\resizebox{\columnwidth}{!}{
\begin{tabular}{@{}lll@{}}
\toprule
Layer                                           & Output           & Activation\\ \midrule
Input                                           & $256\times64\times64$  & None               \\
ResnetBlock  & $256\times64\times64$ & None  \\
ResnetBlock  & $256\times64\times64$ & None  \\
ResnetBlock  & $256\times64\times64$ & None  \\
$128\times3\times3-2-1$ TConv  & $256\times128\times128$ & ReLU  \\
$64\times3\times3-2-1$ TConv  & $64\times256\times256$ & ReLU  \\
ReflectionPad p=3 & $64\times262\times262$ & None \\
$1\times7\times7-1-0$ Conv & $3\times256\times246$ & None \\ \bottomrule
\end{tabular}
}
\end{table}

\end{document}